\documentclass[letterpaper, 10 pt, conference]{ieeeconf}

\usepackage{cite}
\usepackage{graphicx}
\usepackage{framed}
\usepackage{subfig}
\usepackage{indentfirst}
\usepackage[]{algorithm2e}

\usepackage{amsmath, amssymb,amsthm}
\usepackage[font=small]{caption}

\usepackage{comment}
\usepackage{cite}

\IEEEoverridecommandlockouts                
\title{\LARGE \bf 
	Probabilistic Prediction of Interactive Driving Behavior via \\Hierarchical Inverse Reinforcement Learning}
\author{Liting Sun$^{1}$, Wei Zhan$^{1}$, and Masayoshi Tomizuka$^{1}$
	\thanks{*This work was partially supported by the international Chair Drive for All, Foundation MINES ParisTech.}
	\thanks{$^{1}$L. Sun, W. Zhan, and M. Tomizuka are with the Department of Mechanical Engineering, University of California, Berkeley, CA 94720 USA (e-mail: \tt\small litingsun, wzhan, tomizuka@berkeley.edu). }
}

\begin{document}

\maketitle

\begin{abstract}
Autonomous vehicles (AVs) are on the road. To safely and efficiently interact with other road participants, AVs have to accurately predict the behavior of surrounding vehicles and plan accordingly. Such prediction should be probabilistic, to address the uncertainties in human behavior. Such prediction should also be interactive, since the distribution over all possible trajectories of the predicted vehicle depends not only on historical information, but also on future plans of other vehicles that interact with it. To achieve such interaction-aware predictions, we propose a probabilistic prediction approach based on hierarchical inverse reinforcement learning (IRL). First, we explicitly consider the hierarchical trajectory-generation process of human drivers involving both discrete and continuous driving decisions. Based on this, the distribution over all future trajectories of the predicted vehicle is formulated as a mixture of distributions partitioned by the discrete decisions. Then we apply IRL hierarchically to learn the distributions from real human demonstrations. A case study for the ramp-merging driving scenario is provided. The quantitative results show that the proposed approach can accurately predict both the discrete driving decisions such as yield or pass as well as the continuous trajectories.
\end{abstract}

\section{Introduction}
Autonomous vehicles (AVs) are interacting with human on the road. To this end, it needs to predict and reason about possible future behavior of human, and plan its own trajectories accordingly. Wrong prediction can cause either too conservative motions such as unnecessary stops/yielding, or dangerous situations like emergency brakes and unavoidable collisions. Hence, accurate prediction is of crucial importance to enable a safe and efficient autonomous car. 

Starting from as simple as assuming that the other drivers would maintain their current velocities within the planning horizon \cite{kuwata2009real}\cite{liang2012automatic} to as complicated as modelling the probability distributions over all possible trajectories/actions, many prediction approaches have been proposed. In terms of prediction output, they can be categorized into two groups: deterministic prediction \cite{kuwata2009real, liang2012automatic, ziegmann_analysis_2017, Graflearningprediction2014} and probabilistic prediction \cite{lefevre_intention-aware_2013, schreier_integrated_2016, geng_scenario-adaptive_2017}. Both categories can be implemented based on various models such as neural networks (NN) \cite{phillips_generalizable_2017}\cite{hu2018probabilistic}, hidden Markov models (HMM) \cite{dong_intention_2017} and Bayes net \cite{schreier_integrated_2016}. Most of the work, however, formulated the future trajectories/actions as either deterministic functions or conditional probabilities of historical and current scene states. The influence of human drivers' beliefs about the other vehicles future actions are ignored in the prediction framework.

In fact, in interactive driving scenarios, human drivers will actively anticipate and reason about surrounding vehicles' behavior when they decide the next-step actions. This means that besides historical and current scene states, the distribution over all possible trajectories is also influenced by their beliefs about other vehicles' plan.  For example, given the same historical trajectories, a lane-keeping driver might be more probable to decelerate rather than maintaining current speed if the driver thinks another vehicle is about to merge into his lane. Such interaction-aware prediction is incorporated implicitly into a planning framework in \cite{sadigh2016planning}, but the prediction output is deterministic.

\emph{Our insight is that in highly interactive driving scenarios, an accurate prediction should consider not only the time-domain dependency but also the interaction among vehicles. Namely, an autonomous car should anticipate the conditional probability over all possible trajectories of the human driver given not only historical and current states, but also its own future plans.}

To obtain such a probabilistic and interactive prediction of human drivers' behavior, we need to approximate the ``internal incentive'' that drives human to generate specific behavior. Note that human's planning procedure is naturally hierarchical, involving both discrete and continuous driving decisions.  The discrete driving decisions determine the game-theoretic outcomes of interaction such as to yield or to pass, whereas the continuous driving decisions influence details of the resulting trajectories in terms of smoothness, distances to other road participants and higher-order dynamics such as velocities, accelerations and jerks. Hence, to describe the influence of both discrete and continuous decisions, we propose a hierarchical inverse reinforcement learning (IRL) framework in this paper, to learn trajectory-generation process of human from observed demonstrations. Given the autonomous vehicle's future plan, the probability distribution over all future trajectories of the predicted human driver is modelled as a mixture of distributions, partitioned by the discrete driving decisions.

Contributions of this paper are the following two.

\noindent\textbf{A formalism for probabilistic interactive prediction of human driving behavior.} We model the prediction problem from the perspective of a two-agent game by explicitly considering the responses of one agent to another. We formalize the distribution of predicted vehicle's future trajectories as a probability distribution conditioned not only on historical and current information, but also on future plans of the ego vehicle.

\noindent\textbf{A hierarchical IRL framework.} We explicitly consider the hierarchical planning procedure of human drivers, and formulate the influences of both discrete and continuous driving decisions. Via hierarchical IRL, the conditional distribution over all possible future trajectories is expressed as a mixture of probability distributions partitioned by different game-theoretic outcomes of interactive vehicles.

\noindent\textbf{\textit{Related Work in IRL:}} Initially proposed by Kalman \cite{kalman1964linear}, the concept of Inverse Reinforcement Learning is first formulated in \cite{ng2000algorithms}. It aims to infer the reward/cost functions of agents from their observed behavior by assuming that the agents are rational. To deal with uncertainties and noisy observations, Ziebart \textit{et al}.~\cite{ziebart2008maximum} extended the algorithm based on the principle of maximum entropy. It assumes that agents actions/behavior with lower cost are exponentially more probable, and thus an exponential distribution family can be established to approximate the distribution of actions/behavior. Building on this, Levine \textit{et al}.~\cite{levine2012continuous} formulated the continuous IRL algorithm and used it to minic and predict human driving behavior, and Liu \textit{et al}.~\cite{liu2013understanding} used IRL to approximate the decision-making process of taxi drivers and passengers in public transportation. In \cite{shimosaka2015predicting}, Shimosaka \textit{et al}.~predicted human driving behavior in diverse environment by formulating IRL with multiple reward functions. Kretzschmar \textit{et al}.~\cite{kretzschmar2016socially} considered the hierarchical trajectory-generation process of human navigation behavior, but they focus on cooperative agents that share the same reward function instead of interactive agents that have significantly different reward functions.

\section{Problem Statement}
We focus on predicting human drivers' interactive behavior within two vehicles: a host vehicle (denoted by $(\cdot)_H$) and a predicted vehicle (denoted by $(\cdot)_M$). All other in-scene vehicles are treated as surrounding vehicles, denoted by $(\cdot)^i_O$ ($i{=}1,2,{\cdots}, N$ is the index for surrounding vehicles). We use $\xi$ to represent historical vehicle trajectories, and $\hat{\xi}$ for future trajectories\footnote[1]{Note that trajectory is a sequence of states, i.e., $\xi=[x^T_1,x^T_2,\cdots,x^T_L]^T$ where $x_i$ is the vehicle state at $i$-th time step. $L$ is the trajectory length. Depending on the representation, the vehicle state can be different. For instance, it can simply be the positions of vehicles, or it can include velocities, yaw angles, accelerations, etc.}. 

It is obvious that the probability distribution over all possible trajectories of the predicted vehicle depends on his own historical trajectory and those of surrounding vehicles. Mathematically, such time-domain state dependency can be modelled as a conditional probability density function (PDF), as in \cite{lenz2017deep}:
\begin{eqnarray}
p ( \hat{\xi}_M | \xi^{1:N}_{O}, \xi_H, \xi_M).
\label{eq1}
\end{eqnarray}

However, as discussed above, in interactive driving scenarios, the influence of human's beliefs about others' next-step actions cannot be ignored in order to get a good prediction. From this perspective, the probability distribution over all possible trajectories of the predicted vehicle should also be conditioned on potential plans of the host vehicle. Mathematically, such interaction-dependency further refines the conditional PDF in (\ref{eq1}) as 
\begin{eqnarray}
p ( \hat{\xi}_M | \xi^{1:N}_{O}, \xi_H, \xi_M, \hat{\xi}_H).
\label{eq2}
\end{eqnarray}

The key aspect of this formulation is to actively consider the influence of the host vehicle's actions when predicting the future trajectories of the predicted vehicle. In this sense, we are trying to enable the autonomous vehicle to ``think'' in the way that the predicted human driver thinks if he had known the autonomous vehicle's future plan.

\section{Modeling Human Driving Behavior}
\label{sec:human_modeling}
In order to predict human's interactive driving trajectories, we need to model the internal incentive of human that generates his behavior.
\subsection{Probabilistic Hierarchical Trajectory-Generation Process}
As addressed above, the trajectory-generation process of human drivers is naturally probabilistic and hierarchical. It involves both discrete and continuous driving decisions. The discrete driving decisions determine ``rough'' pattern (or homotopy class) of his future trajectory as a game-theoretic result (e.g., to yield or to pass), whereas the continuous driving decisions influence details of the trajectory such as velocities, accelerations and smoothness.

\begin{figure}[h!]
	\centering
	\includegraphics[width=7cm]{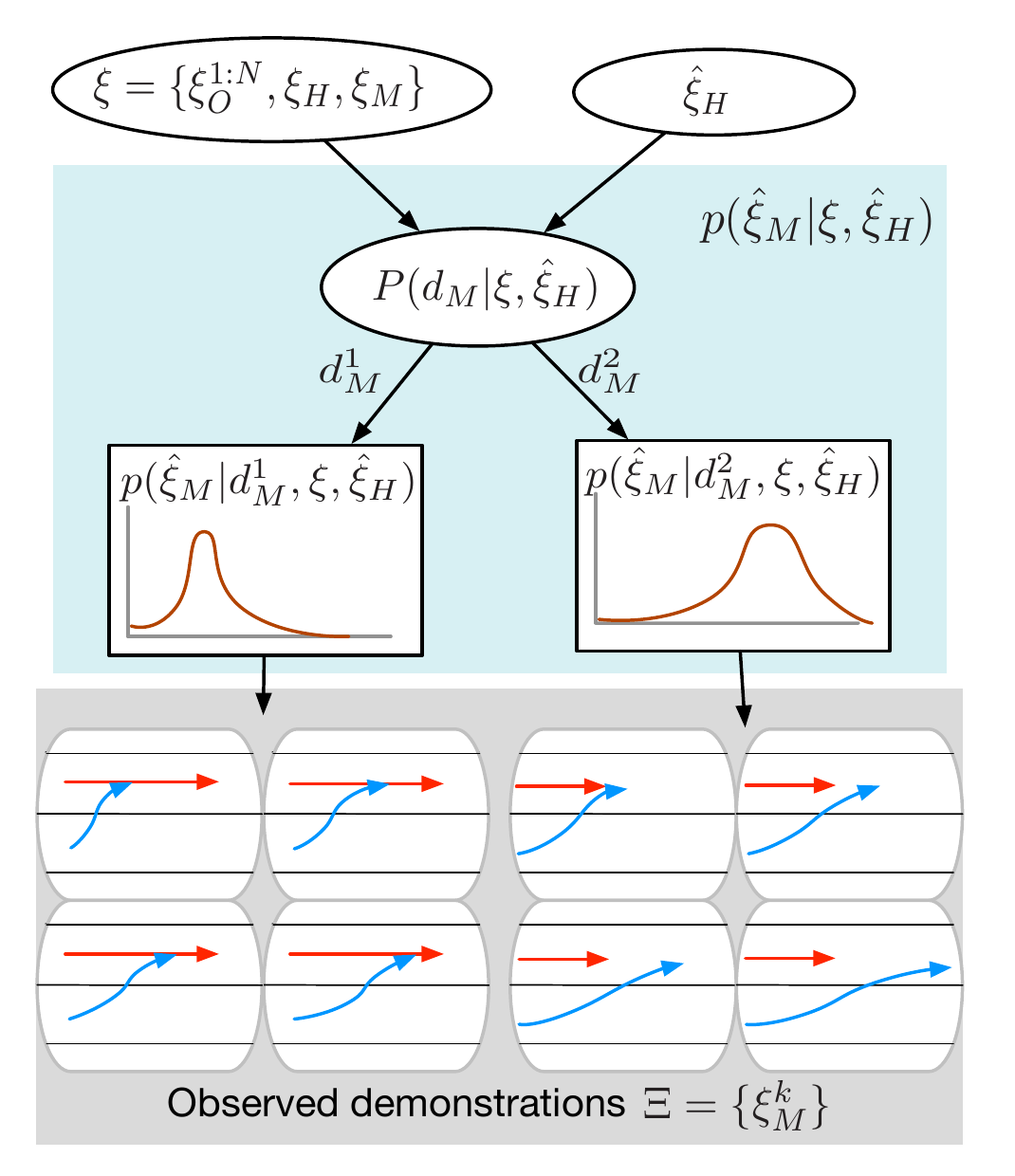}
	\caption{The probabilistic and hierarchical trajectory--generation process for a lane changing scenario. The predicted vehicle (blue) is trying to merge into the lane of the host vehicle (red). Given all observed historical trajectories $\xi{=}\{\xi^{1:N}_O, \xi_H, \xi_M\}$ and his belief about the host vehicle's future trajectory $\hat{\xi}_H$, the trajectory distribution of the predicted vehicle over all the trajectory space is partitioned first by the discrete decisions: merge behind ($d^1_M$) and merge front ($d^2_M$). Under different discrete decisions, the distributions of continuous trajectories can be significantly different, and each of them is represented via a probability distribution model. The observed demonstrations are samples satisfying the distributions.\label{fig:hIRL}}
\end{figure}

Figure \ref{fig:hIRL} illustrates such probabilistic and hierarchical trajectory--generation process for a lane-changing driving scenario. The predicted vehicle (blue) is trying to merge into the lane occupied by the host vehicle (red).  Given all observed historical trajectories $\xi{=}\{\xi^{1:N}_O, \xi_H, \xi_M\}$ and his belief about the host vehicle's future trajectory $\hat{\xi}_H$, he first decides whether to merge behind the host vehicle ($d^1_M$) or merge front ($d^2_M$). Such discrete driving decisions are outcomes of the first-layer probability distribution $P(d_M|\xi,\hat{\xi})$, and partition the space of all possible trajectories into two distinct homotopy classes\footnote[2]{Two trajectories belong to a same homotopy class if they can be continuously transformed to each other without collisions \cite{kretzschmar2016socially}.}, each of them can be described via a second-layer probability distribution $p(\hat{\xi}_M|d_M, \xi, \hat{\xi}_H)$. Note that among different homotopy classes, the distributions of the continuous trajectories can be significantly different since the driving preference under different discrete decisions may be different. For example, a vehicle decided to merge behind might care more about comfort and less about speed, while a vehicle merging front might care about completely the opposite.

Hence, the conditional distribution $p(\hat{\xi}_M|\xi, \hat{\xi}_H)$ in (\ref{eq2}) is formulated as a mixture of distributions, which explicitly captures the influences of both discrete and continuous driving decisions of human drivers:
\begin{equation}
\label{eq:mix_distribution}
p(\hat{\xi}_M|\xi, \hat{\xi}_H)=\sum_{d^i_M\in\mathcal{D}_M}{p(\hat{\xi}_M|d^i_M,\xi,\hat{\xi}_H)P(d^i_M|\xi,\hat{\xi}_H)}
\end{equation}
where $\mathcal{D}_M$ represents the set of all possible discrete decisions for the predicted vehicle.
\subsection{Hierarchical Inverse Reinforcement Learning}
Equation (\ref{eq:mix_distribution}) suggests that in order to model the conditional PDF in (\ref{eq2}) for interactive prediction, we need to model the hierarchical probabilistic models $P(d_M|\xi,\hat{\xi}_H)$ and $p(\hat{\xi}_M|d^i_M,\xi,\hat{\xi}_H)$ for each $d^i_M{\in}\mathcal{D}_M$.

We thus propose to apply inverse reinforcement learning hierarchically to learn all the models from observed demonstrations of human drivers. Based on the principle of maximum entropy \cite{ziebart2008maximum}, we assume that all drivers are exponentially more likely to make decisions (both discrete and continuous) that lead to a lower cost. This introduces an family of exponential distributions depending on the cost functions, and our interest is to find the optimal hierarchical cost functions that ultimately lead to trajectory distributions matching that of the observed trajectories in a given demonstration set $\Xi$.
\subsection{Modeling Continuous Driving Decisions}
Suppose that the demonstration set $\Xi$ is partitioned into $|\mathcal{D}|$ subsets by the discrete decisions $d^i_M{\in}\mathcal{D}$. $\vert\mathcal{D}\vert$ is the dimension of $\mathcal{D}$. Each subset $\Xi_{d^i_M}$ contains trajectories that belong to the same homotopy class. We represent each demonstration in $\Xi_{i}$ by a tuple $(\hat{\xi}_M, d_M^i, \xi, \hat{\xi}_H)$ where $\xi{=}[\xi^{1:N}_O, \xi_H, \xi_M]$ represents all historical information. Since the trajectory space is continuous and demonstrations are noisily local-optimal, we use Continuous Inverse Optimal Control with Locally Optimal Examples\cite{levine2012continuous}.
\subsubsection{Continuous-Space IRL}
Under discrete decision $d_M$, we assume that the cost of each trajectory can be linearly parametrized by a group of selected features $\{\mathbf{f}_{d_M}\}$, i.e., $C(\pmb{\theta}_{d_M}, \hat{\xi}_M, \xi, \hat{\xi}_H)=\mathbf{\pmb\theta}^T_{d_M}\mathbf{f}_{d_M}(\hat{\xi}_M, \xi, \hat{\xi}_H)$, where $\theta_{d_M}$ is the parameter vector to determine the emphasis of each of the features. Then trajectories with higher cost are exponentially less likely based on the principle of maximum entropy :
\begin{equation}
\label{eq:continuous_irl}
P(\hat{\xi}_M|\pmb\theta_{d_M}, d^i_M,\xi,\hat{\xi}_H)\propto e^{-C(\pmb\theta_{d_M}, \hat{\xi}_M, \xi, \hat{\xi}_H)}
\end{equation}
Hence, the log likelihood of the given demonstration subset $\{\Xi_{d_M}\}$ is given by
\begin{eqnarray}
\label{eq:likelihood_Xi}
\log P(\Xi_{d_M}|\pmb\theta_{d_M}){=}\sum_{\hat{\xi}_M\in\Xi_{d_M}}\log\dfrac{e^{-C(\pmb\theta_{d_M}, \hat{\xi}_M, \xi, \hat{\xi}_H)}}{\int e^{-C(\pmb\theta_{d_M}, \tilde{\xi}_M, \xi, \hat{\xi}_H)}d\tilde{\xi}_M}.
\end{eqnarray}
Our goal is to find the optimal $\pmb\theta_{d_M}$ such that the given demonstration set is most likely to happen:
\begin{equation}
\label{eq:goal_max_likelihood}
\pmb\theta_{d_M}^*=\arg\max_{\pmb\theta_{d_M}} P(\Xi_{d_M}|\pmb\theta_{d_M})
\end{equation}
To tackle the normalization factor in (\ref{eq:likelihood_Xi}), we use Laplace approximation as in \cite{levine2012continuous}. Namely, the cost along an arbitrary trajectory $\tilde{\xi}_M$ is approximated by its second-order Taylor expansion around the demonstrated trajectory $\hat{\xi}_M$:
\begin{eqnarray}
\label{eq: Gaussian_approximation}
C(\pmb\theta_{d_M}, \tilde{\xi}_M, \xi, \hat{\xi}_H){\approx}C(\pmb\theta_{d_M}, \hat{\xi}_M, \xi, \hat{\xi}_H)+(\tilde{\xi}_M-\hat{\xi}_M)^T\dfrac{\partial C}{\partial \xi_M}\nonumber\\
+(\tilde{\xi}_M-\hat{\xi}_M)^T\dfrac{\partial^2 C}{\partial \xi_M^2}(\tilde{\xi}_M-\hat{\xi}_M)\qquad\nonumber
\end{eqnarray}
This enables the normalization factor become a Gaussian integral and can be solved analytically. Define $g_{\hat{\xi}_M}(\pmb\theta_{d_M}){=}\frac{\partial C}{\partial \xi_M}|_{\hat{\xi}_M}$ and $H_{\hat{\xi}_M}(\pmb\theta_{d_M}){=}\frac{\partial^2 C}{\partial \xi_M^2}|_{\hat{\xi}_M}$ as, respectively, the gradient and Hessian of the cost along trajectory $\hat{\xi}_M$. Then (\ref{eq:goal_max_likelihood}) is translated to the optimization problem in (\ref{eq:final_goal_likelihood}), which intuitively means the optimal cost function should have small gradient and large positive Hessians along the demonstrated trajectories. For details, one can refer to \cite{levine2012continuous}.
\begin{equation}
\label{eq:final_goal_likelihood}
\min_{\pmb\theta_{d_M}}\sum_{\hat{\xi}_M\in\Xi_{d_M}}g_{\hat{\xi}_M}^{T}(\pmb\theta_{d_M})H_{\hat{\xi}_M}^{-1}(\pmb\theta_{d_M})g_{\hat{\xi}_M}(\pmb\theta_{d_M})-\log\vert H_{\hat{\xi}_M}(\pmb\theta_{d_M})\vert
\end{equation}
\subsubsection{Features}
The features we selected to parametrize the continuous trajectories can be grouped as follows:
\begin{itemize}
\item Speed - The incentive of the human driver to reach a certain speed limit $v_{\lim}$ is captured by the feature
\begin{equation}
\label{eq:velocity_feature}
f_v(\hat{\xi}_M) = \sum_{t=0}^L (v_t-v_{\lim})^2
\end{equation}
$v_t$ is the speed at time $t$ along trajectory $\hat{\xi}_M$ and $L$ is the length of the trajectory.
\item Traffic - In dense traffic environment, human drivers tend to follow the traffic. Hence, we introduce a feature based on the intelligent driver model (IDM) \cite{kesting2010enhanced}
\begin{equation}
\label{eq:traffice_feature}
f_{\text{IDM}}(\hat{\xi}_M) = \sum_{t=0}^L (s_t-s_{t}^{\text{IDM}})^2
\end{equation}
where $s_t$ is the actual spatial headway between the front vehicle and predicted vehicle at time $t$ along trajectory $\hat{\xi}_M$, and $s_{t}^{\text{IDM}}$ is the spatial headway suggested by IDM.
\item Control effort and smoothness - Human drivers typically prefer to drive efficiently and smoothly, avoiding unnecessary accelerations and jerks. To address such preference, we introduce a set of kinematics-related features:
\begin{equation}
\label{eq:kinematics}
f_{\text{acc}}(\hat{\xi}_M) =\sum_{t=0}^L a_t^2, \quad f_{\text{jerk}}(\hat{\xi}_M) =\sum_{t=1}^L \left(\dfrac{a_t-a_{t-1}}{\triangle t} \right)^2
\end{equation}
where $a_t$ represents the acceleration at time $t$ along the trajectory $\hat{\xi}_M$. $\triangle t$ is the sampling time.
\item Clearance to other road participants - Human drivers care about their distances to other road participants when they drive since distance is crucially related to safety. Hence, we introduce a distance-related feature
\begin{equation}
\label{eq:distance_feature}
f_{\text{dist}}(\hat{\xi}_M) = \sum_{t=0}^L\sum_{k=1}^{N{+}1} e^{-\frac{(x_t-x^k_t)^2}{l^2}-\frac{(y_t-y^k_t)^2}{w^2}}
\end{equation}
where $(x_t, y_t)$ and $(x^k_t, y^k_t)$ represent, respectively, the coordinates of the predicted vehicle along $\hat{\xi}_M$  and those of the $k$-th surrounding vehicle. Parameters $l$ and $w$ are the length and width of the predicted vehicle. We use coordinates in Frenet Frame to deal with curvy roads, i.e., $x$ denotes the travelled distance along the road and $y$ is the lateral deviation from the lane center.
\item Goal - This feature describes the short-term goals of human drivers. Typically, goals are determined by the discrete driving decisions. For instance, if a lane-changing vehicle decides to merge in front of a host vehicle on his target lane, he will set his short-term goals to be ahead of the host vehicle. The goal-related feature is given by
\begin{equation}
\label{eq:goal_feature}
f_{g}(\hat{\xi}_M)= \sum_{t=0}^L\Vert(x_t, y_t)-(x_t^g, y_t^g)\Vert^2_2
\end{equation}
\item Courtesy - Most of human drivers view driving as a social behavior, meaning that they not only care about their own cost, but also care about others' cost, particularly when they are merging into others' lanes \cite{Liting2018IROS}. Suppose that the cost of the host vehicle is $C_H(\hat{\xi}_M, \xi, \hat{\xi}_H)$, then to address the influence of courtesy to the interaction, we introduce the feature
\begin{equation}
\label{eq:court_feature}
	f_{\text{court}}(\hat{\xi}_M) = \max\left\{C_H\left(\hat{\xi}_M, \xi, \hat{\xi}_H\right){-}C_H^{\text{default}}, 0\right\}.
\end{equation}
This feature describes the possible extra cost brought by the trajectory $\hat{\xi}_M$ of the predicted vehicle to the host vehicle, compared to the default host vehicle's cost $C_H^{\text{default}}$.  We can learn about $C_H(\cdot)$ also from demonstrations. Details about this will be covered in the case study.
\end{itemize} 

For vehicles in different driving scenarios, or under different discrete driving decisions, the features we used to parametrize their costs are different subsets of the above listed ones. For instance, drivers decided to merge behind or front would set different goals, and drivers with right of way would most likely care less about courtesy than those without.

\textit{Remark I}: Note that all variables $v_t, a_t, \delta_t, x_t$ and $y_t$ in (\ref{eq:velocity_feature})-(\ref{eq:court_feature}) can be expressed as functions of trajectory $\hat{\xi}_M$. For instance, if we define $\hat{\xi}_M=[x_0, y_0, \cdots, x_L, y_L]^T$ where $x$ and $y$ are the coordinates in Frenet Frame, then we can obtain all variables via numerical differentiation. Details are omitted.

\subsection{Modeling Discrete Driving Decisions}
\subsubsection{Features}
Different from the continuous driving decisions that influence the higher-order dynamics of the trajectories, the discrete decisions determine the homotopy of the trajectories. To capture this, we selected the following two features to parametrize the cost function that induces the discrete decisions:
\begin{itemize}
\item Rotation angle - To describe the interactive driving behavior such as overtaking from left or right sides, merging in from front or back, we compute the rotation angle from $(x_t, y_t)$ to $(x_{t,H}, y_{t,H})$ along trajectory $\hat{\xi}_M{\in}\Xi_{d_M}$. Define the angle as $\omega_t$, and then the rotation angle feature is given by
\begin{equation}
\label{eq:feature_angle}
f_{\angle}(d_M)=\sum_{t=0}^L\omega_t
\end{equation}
where $d_M$ is a discrete decision and $L$ is the length of the trajectory $\hat{\xi}_M$.
\item Minimum cost - It is also possible that human drivers make discrete decisions by evaluating the cost of trajectories under each decision, and select the one leading to the minimum-cost trajectory. To address this factor, we consider the feature 
\begin{equation}
\label{eq:feature_mini_cost}
f_{\text{cost}}(d_M)=\min_{\tilde{\xi}_M}\pmb\theta_{{d}_M}^T \mathbf{f}_{d_M}(\tilde{\xi}_M, \xi, \hat{\xi}_H).
\end{equation}
where $\pmb\theta_{{d}_M}$ and $\mathbf{f}_{d_M}$, respectively, represent the learned parameters and selected features for the continuous trajectory distribution under discrete decision $d_M$.
\end{itemize} 

\subsubsection{Discrete-Space IRL} Similarly, we assume that the decisions with lower cost are exponentially more probable. We also assume that the cost function is linearly parametrized by $\pmb\psi$ and feature vector $\mathbf{f}^{\text{d}}=[f_{\angle}, f_{\text{cost}}]^T$, i.e., $C^{\text{d}}(d_M, \psi)=\pmb\psi^T\mathbf{f}^{\text{d}}(d_M)$. Again, our goal is to find the optimal $\pmb\psi^*$ such that the likelihood of the demonstration set $\Xi$ is maximized:
\begin{equation}
\label{eq:maximum_discrete}
\max_{\pmb\psi}P(\Xi|\pmb\psi) = \max_{\pmb\psi}\prod_{d_M\in\Xi}\dfrac{e^{-\pmb\psi^T\mathbf{f}^{\text{d}}(d_M)}}{\sum_{\tilde{d}_M\in\mathcal{D}}e^{-\pmb\psi^T\mathbf{f}^{\text{d}}(\tilde{d}_M)}}
\end{equation}
By taking the log probability and gradient-descent approach, the parameter $\pmb\psi$ is updated via 
\begin{equation}
\label{eq:discrete_psi_update}
\pmb\psi_{s+1} {=} \pmb\psi_s{-}\alpha\left(\frac{1}{N}\sum_{\Xi}\mathbf{f}^{\text{d}}(d_M){-}{\sum}_{\tilde{d}_M}P(\tilde{d}_M|\pmb\psi_s)f^{\text{d}}(\tilde{d}_M)\right)
\end{equation}
\begin{equation}
\label{eq:discrete_psi_update_2}
P(\tilde{d}_M|\pmb\psi_s)=\dfrac{e^{-\pmb\psi_s^T\mathbf{f}^{\text{d}}(\tilde{d}_M)}}{\sum_{\tilde{d}_M\in\mathcal{D}}e^{-\pmb\psi_s^T\mathbf{f}^{\text{d}}(\tilde{d}_M)}}\qquad\qquad\quad
\end{equation}
where $\alpha$ is the step size and $N$ is the number of demonstrated trajectories in set $\Xi$. To estimate the probability $P(\tilde{d}_M|\pmb\psi_s)$, we use sampling-based method. Detailed implementation will be covered later in the case study.

\section{Case Study}
\label{sec:case_study}
In this section, we apply the proposed hierarchical IRL approach to model and predict the interactive human driving behavior in a ramp-merging scenario.
\subsection{Data Collection}
We collect human driving data from the Next Generation SIMulation (NGSIM) dataset \cite{alexiadis_next_2004}. It captures the highway driving behaviors/trajectories by cameras mounted on top of surrounding buildings. The sampling time of the trajectories is $\triangle t{=}0.01$s. We choose 134 ramp-merging trajectories on Interstate 80 (near Emeryville, California), and separated them into two sets: a training set of size 80 (denoted by $\Xi$, i.e., the human demonstrations), and the other 54 trajectories as the test set.

\begin{figure}[h!]
	\centering
	\includegraphics[width=8.8cm]{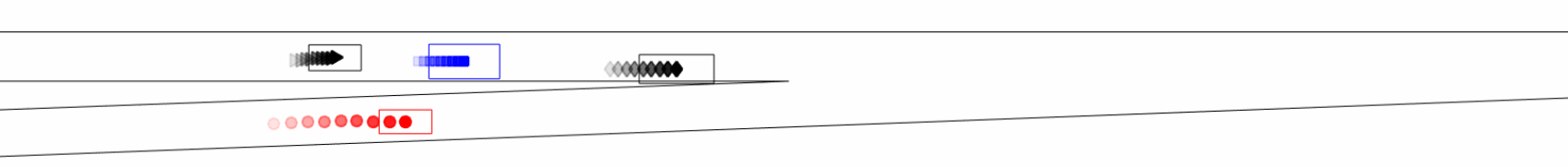}
	\caption{The merging map on Interstate 80 near Emeryville, California. Red: merging vehicle; Blue: lane-keeping vehicle; Black: other surrounding vehicles \label{fig:i80_map}}
\end{figure}

Figure \ref{fig:i80_map} shows the road map (See \cite{alexiadis_next_2004} for detailed geometry information) and an example group of trajectories. There are four cars in scene, one merging vehicle (red), one lane-keeping vehicle (blue) and two surrounding vehicles (black), with one ahead of the blue car and the other behind. Our interest focuses on the interactive driving behavior of both the merging vehicle and the lane-keeping vehicle.
\subsection{Driving Decisions and Feature Selection}
We use the same hierarchical IRL approach to model the conditional probability distributions for both the merging vehicle and the lane-keeping vehicle. 
\subsubsection{Driving Decisions}
In the ramp-merging scenario, the driving decisions are listed as in Table \ref{tab:decisions}. As mentioned above, $\mathbf{x}{=}[x_1, \cdots, x_L]^T$ and $\mathbf{y}{=}[y_1, \cdots, y_L]^T$ are, respectively, the coordinate vectors in Frenet Frame along the longitudinal and lateral directions. $L$ is set to be $50$, i.e., in each demonstration, $5$s trajectories are collected.
\begin{table}[th!]
	\centering
	\begin{tabular}{|c|c|c|}
		\hline 
		&  Discrete Decisions & Continuous Decisions    \tabularnewline
		\hline 
		\hline
		merging-in& $\mathcal{D}=$ & trajectory  \tabularnewline
		vehicle & $\{\text{merge front, merge back}\}$& $\xi{=}[x_1, y_1, \cdots, x_L, y_L ]^T$  \tabularnewline
		\hline 
		lane-keeping&$\mathcal{D}=$& trajectory \tabularnewline
		vehicle & $\{\text{yield, pass}\}$& $\xi{=}[x_1, y_1, \cdots, x_L, y_L ]^T$  \tabularnewline
		\hline
	\end{tabular}
	\caption{Driving decisions for the interactive vehicles}	
	\label{tab:decisions}
\end{table}
\subsubsection{Feature Selection}
Since the right of way for the merging vehicle and the lane-keeping vehicles are different, we define different features for them. 

For the lane-keeping vehicle, the feature vectors related to the continuous driving decisions are as follows:
\begin{itemize}
	\item Yield: $\mathbf{f}_{\text{yield}}{=}[f_v, f_{\text{acc}}, f_{\text{jerk}}, f_{\text{dist}}, f_{g}]^T$. We exclude the feature $f_{\text{IDM}}$ because once the lane-keeping driver decides to yield to the merging vehicle, it is very likely that he cares more about the relative positions to the merging vehicle instead of the heading space to the front vehicle when he plans his continuous trajectories. The goal position in $f_g$ is set to be $[x_\text{current lane center}, y_{t, \text{merging vehicle}}{-}s_0]$, i.e., $s_0$ behind the merging vehicle along longitudinal direction.
	\item Pass: $\mathbf{f}_{\text{pass}}{=}[f_v, f_{\text{IDM}}, f_{\text{acc}}, f_{\text{jerk}}, f_{\text{dist}}, f_g]^T$. In this case, the goal position in $f_g$ is set to be ahead of the merging vehicle along longitudinal direction, i.e., $[x_\text{current lane center}, y_{t, \text{merging vehicle}}{+}s_0]$. Also, if the driver decides to pass, it is more probable that the heading space to the front vehicle will influence the distribution of his continuous trajectories.
\end{itemize}

For the merging vehicle, the feature vectors for the continuous driving models are:
\begin{itemize}
\item Merge back: $\mathbf{f}_{\text{back}}{=}[f_v, f_{\text{acc}}, f_{\text{jerk}}, f_{\text{dist}}, f_{g}]^T$. The goal position is set to be $[x_\text{target lane center}, y_{t, \text{on-lane vehicle}}{-}s_0]$.
\item Merge front: $\mathbf{f}_{\text{front}}{=}[f_v, f_{\text{IDM}}, f_{\text{acc}}, f_{\text{jerk}}, f_{\text{dist}}, f_{\text{court}}, f_{g}]^T$. Once the merging driver decides to merge in front of the lane-keeping vehicle, his heading space to the front lane-keeping vehicle is crucial to the distribution of his possible trajectories. Hence, we include feature $f_{\text{IDM}}$. Moreover, to respect the right of way of the lane-keeping vehicle, merging drivers might care about the extra cost they bring to the lane-keeping vehicle and prefer trajectories that induce less extra cost. To capture such effect, we add $f_{\text{court}}$. Given demonstrated trajectory group $\pmb\xi{=}[\xi_{\text{merging}}, \xi_{\text{surroudings}}, \xi_{\text{lane-keeping}}]$, $f_{\text{court}}$ is computed via (\ref{eq:court_feature}), i.e., 
\begin{eqnarray}
\label{eq:court_computation}
f_{\text{court}}(\xi_{\text{merging}})=C_{\text{lane-keeping}}(\pmb\theta_{\text{yield}}, \pmb\xi)-C_{\text{lane-keeping}}^{\text{default}}
\end{eqnarray}
The cost function $C_{\text{lane-keeping}}(\pmb\theta_{\text{yield}}){=}\pmb\theta_{\text{yield}}^T\mathbf{f}_{\text{yield}}$ is learned with the features selected above. Regarding to $C_H^{\text{default}}$, we assume that the lane-keeping vehicle is by default following IDM.
\end{itemize}

\subsection{Implementation Details and Training Performance}
\begin{figure}[h!]
	\centering
	\includegraphics[width=8.6cm]{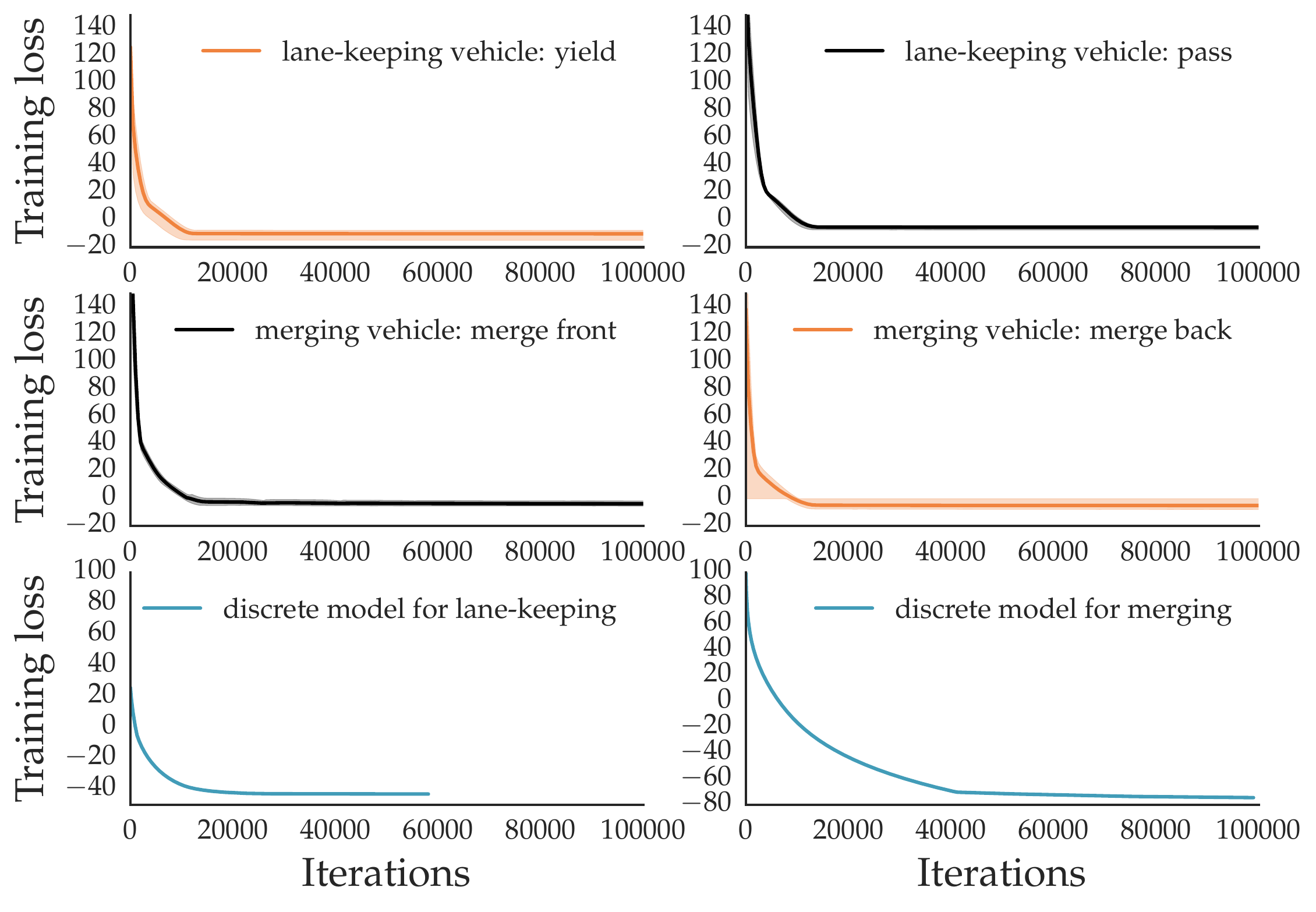}
	\caption{The training curves of both the lane-keeping vehicle and the merging vehicle under different discrete driving decisions \label{fig:training_curve}}
\end{figure}

We use Tensorflow to implement the hierarchical IRL algorithm. Figure \ref{fig:training_curve} gives the training curves regarding to both the continuous and discrete driving decisions. Due to the hierarchical structure, we first learn all four continuous distribution models for both the merging vehicle and the lane-keeping vehicle under different discrete decisions. We randomly sample subsets of trajectories from the training set and perform multiple trains. As seen from Fig.~\ref{fig:training_curve}, the parameters in each continuous model converge quite consistently with small variance. 

With the converged parameter vectors $\pmb\theta_{\text{yield}}$, $\pmb\theta_{\text{pass}}$, $\pmb\theta_{\text{front}}$ and $\pmb\theta_{\text{back}}$, the discrete feature vectors $\mathbf{f}^d$s are then calculated and thus the optimal parameter vectors $\pmb\psi$s are learned via discrete IRL. To efficiently sample continuous trajectories under different discrete decisions, we first obtain the most-likely trajectories by optimizing the learned cost functions under each decision and then randomize them with additive Gaussian noise. The training curves (blue) are also shown in Fig.~\ref{fig:training_curve}.
\subsection{Test Results}

Once $\pmb\theta_{\text{yield}}$, $\pmb\theta_{\text{pass}}$, $\pmb\theta_{\text{front}}$, $\pmb\theta_{\text{back}}$ and $\pmb\psi_{\text{merging}}$, $\pmb\psi_{\text{lane-keeping}}$ are acquired, we can obtain the conditional PDF defined in (\ref{eq2}) via (\ref{eq:mix_distribution}). With this PDF, probabilistic and interactive prediction of human drivers' behavior can be obtained.  The prediction horizon is 3s, i.e., 30 points with a sampling period of 0.1s.
\subsubsection{Accuracy of the probabilistic prediction of discrete decisions}
To measure the accuracy of the probabilistic prediction of discrete decisions, we extract $N{=}2000$ short trajectories in the test set with a horizon length of $10$ from the 54 long trajectories. For each short trajectory, starting from the same initial condition, we sample $M{=}4$ trajectories with different motion patterns (discrete decisions) in the spatiotemporal domain, and one of them is set the same as the ground truth. Hence, the ground truth probabilities for all trajectories are either 0 or 1.

\textbf{Metric:} We adopt a fatality-aware Brier metric \cite{Wei2018ITSC} to evaluate the prediction accuracy. The fatality-aware metric refines the Brier score  \cite{brier1950verification} by formulating the prediction errors in three different aspects: ground-truth accuracy ($\mathcal{G}$) measuring the prediction error of the ground-truth motion pattern ($\Xi_g$), conservatism accuracy ($\mathcal{C}$) measuring the false alarm of aggressive motion patterns ($\Xi_a$), and non-defensiveness accuracy ($\mathcal{D}$) measuring the miss detection of dangerous motion patterns ($\Xi_d$). Let $P_{i,j}$ be the predicted probability of motion pattern $j$ for $i$-th test example ($j{=}1,2,\cdots,M$ and $i{=}1,2,\cdots, N$) , then the three scores and the overall score are given by
\begin{eqnarray}
\label{eq:new_score_separated}
\mathcal{G} &=& \dfrac{1}{|\Xi_{g}|}\sum_{(i,j)\in\Xi_{d}}\left(P_{i,j} - 1\right)^2,\\
\mathcal{C}&=& \dfrac{1}{|\Xi_{a}|}\sum_{(i,j)\in\Xi_{a}}W_{c}(i,j)\left(P_{i,j} - 0\right)^2,\\
\mathcal{D}&=& \dfrac{1}{|\Xi_{d}|}\sum_{(i,j)\in\Xi_{d}}W_{d}(i,j)\left(P_{i,j} - 0\right)^2,\\
\mathcal{B}_c &=& \mathcal{G}+\mathcal{C}+\mathcal{D}.
\end{eqnarray}  
$W_c(i,j)$ and $W_d(i,j)$ are, respectively, the weights penalizing the conservatism and non-defensiveness of the motion pattern $(i,j)$. For more details, one can refer to \cite{Wei2018ITSC}.

\textbf{Comparison:} We compare the prediction among three different approaches: the proposed hierarchical IRL method, a neural-network (NN) based method \cite{bishop1994mixture} and a hidden markov models (HMM) based method \cite{HMM}.

\textbf{Results:} The scores for all three methods are shown in Table \ref{tab:score}. We can see that the proposed method can yield better overall prediction performance than HMM and NN based methods, particularly in terms of conservatism and non-defensiveness. This means that the prediction generated by the proposed method has similar criticality as the ground truth in the interaction process.
\begin{table}[th!]
	\centering
	\begin{tabular}{|c|c|c|c|}
		\hline 
		&  HMM & NN & Hierarchical IRL   \tabularnewline
		\hline 
		\hline
		$\mathcal{G}$ & 0.0701  & 0.0563  &  0.1117  \tabularnewline
		\hline
		$\mathcal{C}$ & 0.0476  & 0.0493  &  0.0178  \tabularnewline
		\hline
		$\mathcal{D}$ & 0.1356  & 0.1303  &  0.0698  \tabularnewline
		\hline
		$\mathcal{B}_c=\mathcal{G}+\mathcal{C}+\mathcal{D}$ & 0.2551  & 0.2361  &  0.2053  \tabularnewline
		\hline
	\end{tabular}
	\caption{Scores of different probabilistic prediction approaches}	
	\label{tab:score}
\end{table}
\subsubsection{Accuracy of the probabilistic prediction of continuous trajectories}
\begin{figure}[h!]
	\centering
	\includegraphics[width=8.6cm]{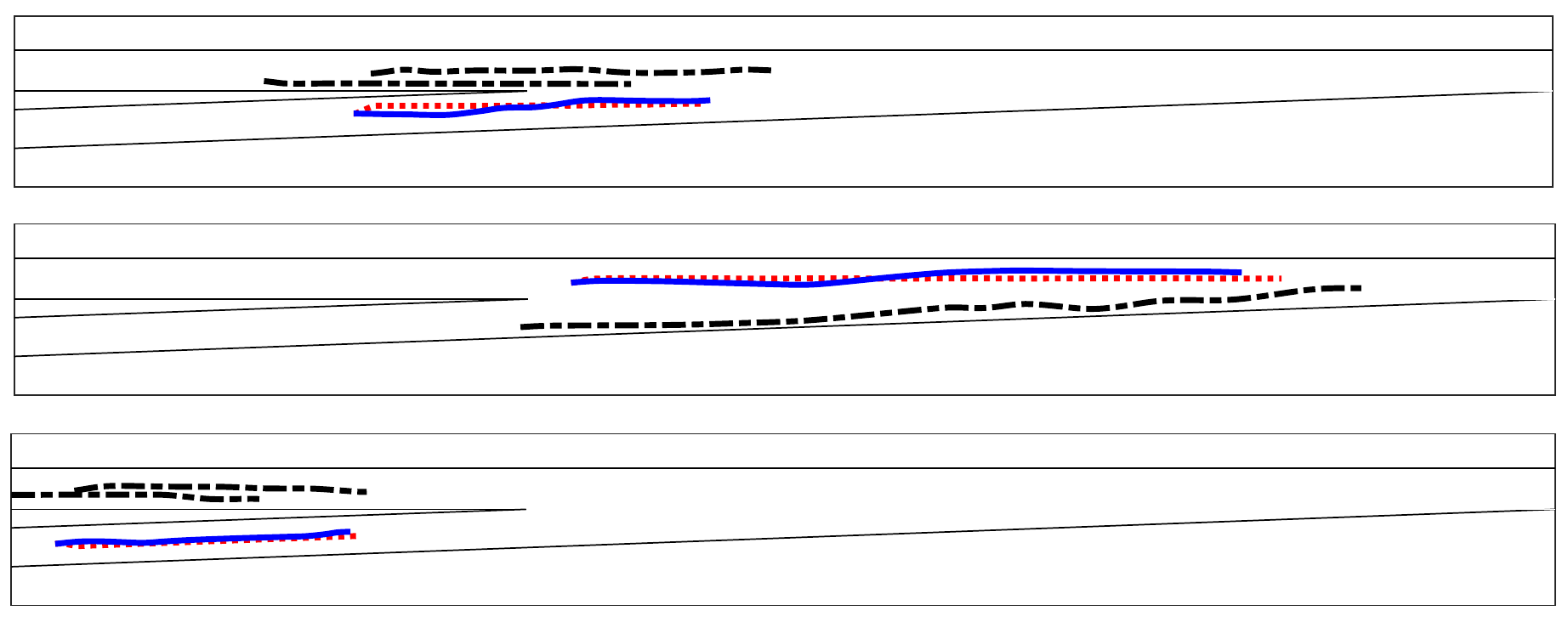}
	\caption{Three illustrative examples of the predicted most probable trajectories (red dotted line) compared with the ground truth trajectories (blue solid line). Thick black dash-dot lines represent the trajectories of other vehicles except for the predicted one.\label{fig:example_predicted_trajectories}}
\end{figure}

In this test, we generate the most probable trajectories under different discrete driving decisions by solving a finite horizon Model Predictive Control (MPC) problem using the above learned continuous cost functions. We show three illustrative examples in Fig.~\ref{fig:example_predicted_trajectories}. The red dotted lines and blue solid lines represent, respectively, the predicted most-likely trajectories and the ground truth trajectories. The thick black dash-dot lines are trajectories of other vehicles. We can see that the predicted trajectories are very close to the ground truth ones. 

\textbf{Metric:} We also adopt Mean Euclidean Distance (MED) \cite{quehl_how_2017} to quantitatively evaluate the accuracy of the prediction for continuous trajectories. Given the ground truth trajectory $\xi_{\text{ground}}{=}[x_{g,1}, y_{g,1}, {\cdots}, x_{g,L}, y_{g,L}]^T$ and predicted trajectory $\xi_{\text{prediction}}{=}[x_{p,1}, y_{p,1}, {\cdots}, x_{p,L}, y_{p,L}]^T$ of same length $L$ and same sampling time $\triangle T$, the trajectory similarity is calculated as follows:
\begin{equation}
\mathcal{S}_{\text{MED}} = \dfrac{1}{L}\sum_{i{=}1}^L\Vert\left[x_{p,i}, y_{p,i}\right]^T - \left[x_{g,i}, y_{g,i}\right]^T\Vert_2
\end{equation}

\textbf{Results:} We test on $20$ long trajectories in the test set, and results are summarized in Table \ref{tab:MED}: the proposed hierarchical IRL method can achieve trajectory prediction with a mean MED of 0.6172m with standard deviation of 0.2473m.
\begin{table}[th!]
	\centering
	\begin{tabular}{|c|c|c|c|c|}
		\hline 
		& Mean (m)  & Max (m) & Min (m) & Std (m)   \tabularnewline
		\hline 
		\hline
		MED & 0.6172 & 1.0146  & 0.3644 &  0.2473 \tabularnewline
		\hline
	\end{tabular}
	\caption{Trajectory similarities in terms of MED}	
	\label{tab:MED}
\end{table}
\section{Conclusion}
In this work, we have proposed a probabilistic and interactive prediction approach via hierarchical inverse reinforcement learning. Instead of directly making predictions based on historical information, we formulate the prediction problem from a game--theoretic view: the distribution of future trajectories of the predicted vehicle strongly depends on the future plans of the host vehicle. To address such interactive behavior, we design a hierarchical inverse reinforcement learning method to capture the hierarchical trajectory--generation process of human drivers. Influences of both discrete and continuous driving decisions are explicitly modelled. We also have applied the proposed method on a ramp--merging driving scenario. The quantitative results verified the effectiveness of the proposed approach in terms of accurate prediction to both discrete decisions and continuous trajectories.

\section*{Acknowledgment}

We thank Yeping Hu and Jiachen Li for their assistance with the implementation of the neural network (NN) and hidden Markov model (HMM) based prediction approaches in Section IV-D.
\bibliographystyle{IEEEtran}
\bibliography{ITSC2018.bib}

\end{document}